\documentclass[letterpaper, 10 pt, conference]{ieeeconf}  

\usepackage{amssymb}
\usepackage{lipsum}
\usepackage{graphicx}
\usepackage{xcolor}
\usepackage{float}
\usepackage{amsmath}
\usepackage{algpseudocode}
\usepackage{algorithm}
\usepackage{wrapfig}
\usepackage{bm}
\usepackage{multirow}
\usepackage{booktabs}
\usepackage{siunitx}
\usepackage{tabularx}
\usepackage{makecell}
\usepackage[
    backend=biber,
    style=ieee,
    sorting=none,
    sortcites=true,
    doi=false,
    url=false,
    eprint=true,
    maxnames=4,
    giveninits=true
]{biblatex}
\usepackage{authblk}

\DeclareFieldFormat[article]{volume}{vol.\ #1}
\DeclareFieldFormat[article]{number}{no.\ #1}

\AtEveryBibitem{%
  \ifentrytype{article}{%
    \clearfield{month}%
  }{}%
}

\AtEveryBibitem{%
  \clearfield{eventtitle}%
}

\AtEveryBibitem{\clearfield{eprintclass}}

\makeatletter
\renewcommand\AB@affilsepx{  \protect\Affilfont}
\makeatother

\newcommand{\richa}[1]{\color{black}{#1}}
\newcommand{\jason}[1]{\color{black}{#1}}

\newcommand{\cyl}[1]{{\color{black}#1}}

\title{\LARGE \bf
Feasibility-Guided Planning over Multi-Specialized Locomotion Policies
}

\author[1*]{Ying-Sheng Luo}
\author[1*]{Lu-Ching Wang}
\author[1*]{Hanjaya Mandala}
\author[1*]{Yu-Lun Chou}
\author[1*]{Guilherme Christmann}
\author[2]{\authorcr Yu-Chung Chen}
\author[2]{Yung-Shun Chan}
\author[2]{Chun-Yi Lee}
\author[1]{Wei-Chao Chen}
\affil[1]{Inventec Corporation}
\affil[2]{National Taiwan University}

\begin{document}

\maketitle

\begingroup\renewcommand\thefootnote{*}
\footnotetext{Joint first authors.}
\endgroup

\thispagestyle{empty}
\pagestyle{empty}

\begin{abstract}

Planning over unstructured terrain presents a significant challenge in the field of legged robotics. Although recent works in reinforcement learning have yielded various locomotion strategies, planning over multiple experts remains a complex issue. Existing approaches encounter several constraints: traditional planners are unable to integrate skill-specific policies, whereas hierarchical learning frameworks often lose interpretability and require retraining whenever new policies are added. In this paper, we propose a feasibility-guided planning framework that successfully incorporates multiple terrain-specific policies. Each policy is paired with a \textit{Feasibility-Net}, which learned to predict feasibility tensors based on the local elevation maps and task vectors. This integration allows classical planning algorithms to derive optimal paths. Through both simulated and real-world experiments, we demonstrate that our method efficiently generates reliable plans across diverse and challenging terrains, while consistently aligning with the capabilities of the underlying policies.

\end{abstract}

\section{Introduction} 
\label{sec:introduction}


\cyl{Effective planning constitutes the foundation of robust robotic navigation on unstructured terrain and fundamentally depends on the underlying locomotion capabilities. Recent developments in both model-predictive control (MPC) ~\cite{dicarloDynamicLocomotionMIT2018a} and model-free reinforcement learning (RL)~\cite{leeLearningQuadrupedalLocomotion2020a} approaches have enabled modern robots to achieve exceptional performance in traversing challenging and unstructured terrain. Enhanced locomotion capabilities expand the accessible planning space through novel regions and enable robots to navigate more effectively across greater distances. However, with the increasing diversity of locomotion methods and skill specialization, the planning space becomes considerably complex. This complexity places substantial demands on planning methodologies, which must reason about the capabilities and limitations of available locomotion skills to generate valid and optimal trajectories. The complex relationship between locomotion diversity and planning complexity further highlights the critical importance of sophisticated planning frameworks that can effectively integrate and coordinate multiple locomotion modalities. These challenges thus motivate the need for innovative planning strategies that can effectively handle complex scenarios while maintaining path optimality across diverse locomotion capabilities.}

\begin{figure}[t!]
    \centering
    \includegraphics[width=0.99\linewidth]{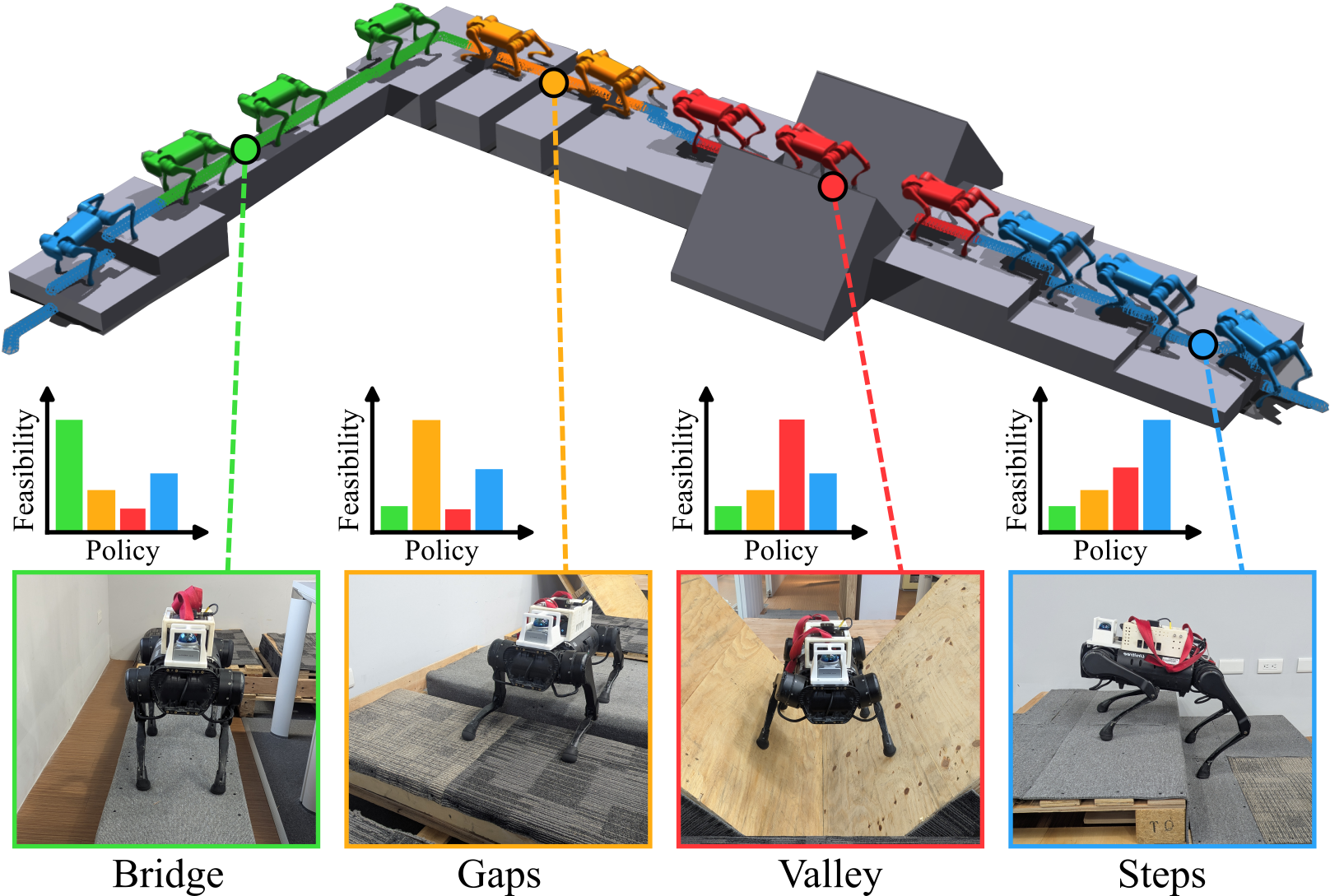}
    \caption{
   \cyl{Feasibility-guided planning enables optimal path selection \& policy switching over mixed terrain via policy-specific feasibility representations.}
    }
    \label{fig:simulation_results}
\end{figure}

\cyl{Traditional planning approaches~\cite{macenskiMarathon2Navigation2020a} attempt to address this complexity through classical robotics navigation methods that rely fundamentally on the elevation data transformed into occupancy grids and costmaps before being solved through graph search algorithms~\cite{macenskiMarathon2Navigation2020a}. These methods require high-quality costmaps that demand extensive empirical tuning within the deployment environment. Classical approaches then establish thresholds to classify cells as obstacles or free space based on local features such as height and surface roughness. Nevertheless, this thresholding process becomes increasingly laborious and imprecise when multiple adaptable policies must be considered. Moreover, thresholding approaches fundamentally lack directional awareness, as traversal from a taller obstacle to a shorter one may be feasible while the reverse direction remains infeasible. To overcome these classical limitations, hierarchical planners trained on top of learned skills have emerged as a common strategy. These learning-based planners can reason about available skills through experience gained in simulated~\cite{hoellerANYmalParkourLearning2024} or real-world rollouts~\cite{weerakoonVAPORLeggedRobot2024}. Nevertheless, such enhanced reasoning capability comes at the expense of decision-making interpretability. The selection rationale for specific skills generally remains opaque to system designers and robot operators. Furthermore, the addition of new skills necessitates complete retraining of the entire system, which limits the scalability and adaptability of such approaches.}

\cyl{While classical methods suffer from interpretability and adaptability issues, RL-based planning approaches present their own challenges~\cite{hendersonDeepReinforcementLearning2018a}. The core difficulty lies in addressing problem sets that require diverse and distinct locomotion capabilities, as training a single policy to handle comprehensive terrain variations often exceeds practical tractability limits.
To manage this complexity, skills are typically compartmentalized into distinct portions of the problem space and subsequently integrated through methods such as mixture-of-experts architectures~\cite{wonScalableApproachControl2020} or hierarchical RL frameworks \cite{pengMCPLearningComposable2019a}. The independent skills can then be sequentially coordinated~\cite{konidarisSkillDiscoveryContinuous2009} through various coordination mechanisms, including timing switch identification~\cite{soesenoTransitionMotionTensor2021, christmannExpandingVersatilityAgile2023a}, continuous modulation techniques~\cite{christmannExpertComposerPolicy2024a}, or specialized gait switching methodologies~\cite{bledtMITCheetah32018, humphreysBioInspiredGaitTransitions2023}.
Despite these coordination strategies, RL-based approaches fundamentally require the planner to accommodate multiple adaptable policies while simultaneously evaluating their respective capabilities and limitations to achieve optimal skill selection and path generation. Such multi-policy coordination scenarios introduce significant overhead and can raise challenges in scalability as the number of locomotion skills increases.}

\cyl{In light of these fundamental limitations across existing planning paradigms, we propose a feasibility-guided planning methodology that addresses the core challenge of coordinating multiple specialized locomotion policies for navigation across heterogeneous terrain, as shown in Fig.~\ref{fig:simulation_results}. Rather than relying on binary occupancy classifications or opaque learned planners, our approach transforms elevation maps into interpretable, policy-specific feasibility representations that preserve the transparency of classical methods while harnessing the capabilities of modern locomotion skills. The key insight underlying our methodology is that effective multi-policy planning requires explicit modeling of each policy's terrain-specific capabilities. We achieve this through \textit{Feasibility-Net} models that are jointly trained with their corresponding locomotion policies to estimate both traversal feasibility and distributional characteristics of training terrain. These models enable us to generate terrain familiarity-weighted feasibility values across elevation maps using a sliding window approach. The resulting representations are inherently grounded in the actual capabilities of their paired policies, enabling traditional graph search algorithms to plan over fused feasibility tensors while maintaining full interpretability of planning decisions. Critically, this architecture supports seamless integration of new policies without retraining. Our contributions are summarized as follows:}


\cyl{\begin{itemize} 
    \item A novel per-policy feasibility estimation framework that jointly learns locomotion-specific traversal predictions and terrain distribution models, thereby enabling robust assessment across diverse and unfamiliar environments.

    \item A unified training paradigm that seamlessly integrates RL policy optimization with supervised feasibility learning, eliminating the need for separate data generation pipelines while maintaining the training efficiency.

    \item A directional feasibility tensor representation that transforms elevation maps into plannable cost functions, while supporting dynamic policy selection and interpretable planning without high-level controller training.
\end{itemize}}

\section{Related Work}

\textbf{Learning-Based Traversability Prediction.} 
\cyl{Learning-based traversability prediction has emerged as a critical component for autonomous navigation in unstructured environments~\cite{zhuLearningBasedTraversabilityCostmap2025, giganteMachineLearningBasedApproach2023, gasparinoWayFASTERSelfSupervisedTraversability2024}. These approaches leverage multi-modal sensor data to model terrain navigability and demonstrate adaptive capabilities when encountering novel environmental conditions~\cite{gasparinoWayFASTNavigationPredictive2022a, sevastopoulosSurveyTraversabilityEstimation2022a}. The complexity of legged locomotion dynamics poses additional challenges for traversability assessment compared to wheeled systems. Contemporary methods address this complexity through the fusion of visual and proprioceptive information for terrain-aware locomotion \cite{leeLearningQuadrupedalLocomotion2020a, zhangTraversabilityAwareLeggedNavigation2024, chengQuadrupedRobotTraversing2024}, while risk-aware prediction frameworks aim to improve reliability~\cite{ganEnergyBasedLeggedRobots2022a, ohTRIPTerrainTraversability2024a, wellhausenRoughTerrainNavigation2021}. Nevertheless, generalization across diverse terrain types and robustness to sensor noise remain persistent challenges~\cite{haLearningbasedLeggedLocomotion2025, haddelerTraversabilityAnalysisVision2022a}. While these approaches provide valuable terrain assessment capabilities, they typically produce single traversability estimates that do not account for the varying capabilities of different locomotion policies. Our approach addresses this limitation by grounding feasibility predictions in the specific capabilities of individual locomotion policies, enabling explicit multi-policy planning across heterogeneous terrain configurations.}

\begin{figure*}[t]
    \vspace{0.2cm}
    \centering
    \includegraphics[width=0.99\linewidth]{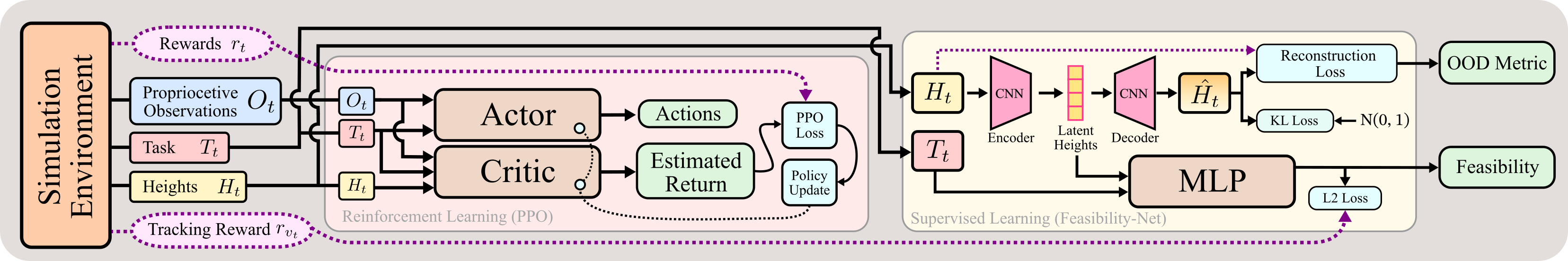}
    \caption{
    \cyl{Overview of the feasibility-aware planning framework. Locomotion policies and \textit{Feasibility-Net} models are jointly trained using shared environment rollouts, where the feasibility models learn to predict velocity tracking performance and terrain distributions simultaneously with policy optimization.}
    }
    \label{fig:system-diagram}
\end{figure*}

\textbf{Skill Chaining/Hierarchical RL.}
\cyl{Hierarchical RL and skill chaining methodologies enable autonomous agents to address long-horizon tasks through systematic decomposition into reusable behavioral components \cite{huangDiscoveringExploitingSkills2024, linHierarchicalLearningRobot1993}. Traditional hierarchical approaches employ high-level controllers to orchestrate the activation of specialized lower-level skills \cite{pateriaHierarchicalReinforcementLearning2021, evansCreatingMultiLevelSkill2023}. These frameworks have demonstrated success in legged locomotion applications \cite{pengDeepLocoDynamicLocomotion2017, jainHierarchicalReinforcementLearning2019, liLearningGeneralizableLocomotion2020, leeLearningRobustAutonomous2024}, particularly through the integration of navigation planners with learned locomotion policies \cite{rudinLearningWalkMinutes2022a}. However, hierarchical approaches introduce fundamental trade-offs between capability and interpretability. The learned high-level controllers typically operate as black-box systems, making it difficult to understand policy selection rationales or to integrate new skills without extensive retraining. Furthermore, these methods often require careful reward engineering and hyperparameter tuning for the hierarchical structure itself. Our methodology circumvents these limitations by employing interpretable graph search algorithms over learned feasibility representations, eliminating the need for high-level controller training while maintaining transparency in policy selection decisions.}

\section{Methodology}

    \cyl{Our approach addresses the multi-policy planning challenge through a unified framework that integrates locomotion policy learning with feasibility estimation, as illustrated in Fig.~\ref{fig:system-diagram}. The framework is motivated by the need to coordinate multiple specialized locomotion policies while maintaining interpretability and avoiding the limitations of hierarchical controllers. The system processes three key inputs from the simulation environment: \textit{proprioceptive observations for policy control}, \textit{local height maps for terrain assessment}, and \textit{task vectors encoding velocity commands}. The core innovation lies in the joint training paradigm where locomotion policies and their corresponding \textit{Feasibility-Net} models are optimized simultaneously through shared environment interactions. While the locomotion policy learns via standard PPO actor-critic architecture \cite{schulmanProximalPolicyOptimization2017a}, the \textit{Feasibility-Net} employs supervised learning to predict velocity tracking rewards based on the same environmental data. The \textit{Feasibility-Net} architecture incorporates a variational autoencoder that models terrain height distributions alongside a multilayer perceptron (MLP) head that generates feasibility predictions, enabling robust performance assessment under distribution shift conditions through out-of-distribution (OOD) detection. This joint training enables each policy to develop terrain specialization while its paired feasibility model learns terrain-specific performance predictions that are inherently grounded in the policy's actual capabilities. During deployment, the framework transforms elevation maps into directional feasibility tensors that encode predicted policy performance across different movement directions, which are subsequently fused to create unified cost functions compatible with standard graph search algorithms. This design enables optimal path discovery with full transparency in policy selection decisions while supporting seamless integration of new policies without the need for retraining.}


\subsection{\cyl{Learning Policy-Specific Feasibility Models}
}

\cyl{We next detail the architecture and training methodology of \textit{Feasibility-Net}, which serves as the core mechanism that enables each locomotion policy to develop terrain-specific performance predictions while maintaining robust generalization capabilities. \textit{Feasibility-Net} takes two primary inputs: (1) a task vector encoding velocity commands, and (2) a local heightmap sampled from the robot's immediate vicinity. The network architecture employs an MLP head that outputs normalized feasibility scores representing the expected velocity tracking performance. We define feasibility as the predicted velocity tracking reward, which is learned through supervised regression with $L2$ loss minimization. The feasibility scores are bounded between $0.0$ and $1.0$, where $0.0$ indicates complete tracking failure and $1.0$ represents perfect velocity following. This formulation integrates seamlessly into standard RL pipelines by utilizing identical rollout data already collected during policy training. The velocity tracking reward function is formally defined as:}
\begin{align}
        r_v &= \exp\left(-\|v - \hat{v}\|_2 / \sigma\right),
    \end{align}
\cyl{where $v$ represents the agent's root velocity vector, $\hat{v}$ denotes the velocity command from the task vector, and $\sigma = 0.25$ serves as a scaling parameter for controlling sensitivity.}

\cyl{A critical challenge in deploying specialized locomotion policies across diverse environments is the inevitable presence of distribution shift between training and deployment terrains. Feasibility predictions become unreliable when the model encounters terrain features that differ significantly from the training distribution. To address this fundamental limitation, we augment the \textit{Feasibility-Net} with a variational autoencoder branch that explicitly models the distribution of training heightmaps. The VAE reconstruction error serves as a metric for OOD detection, where heightmaps exhibiting high reconstruction loss indicate terrain configurations that deviate substantially from the training distribution. This mechanism enables the system to modulate feasibility confidence based on terrain familiarity, thereby improving robustness during deployment.
The complete \textit{Feasibility-Net} optimization combines both feasibility prediction and distribution modeling through the following joint loss function:}

\begin{align}
        \mathcal{L}_{\text{feas}} &= \left\| \hat{f} - r_{v} \right\|_2^2 + \alpha \, \mathcal{L}_{\text{VAE}}, \\
        \mathcal{L}_{\text{VAE}} &= \mathcal{L}_{\text{recon}}(\hm{H}_{x,y}) + \beta \mathcal{L}_{\text{KL}}(\hm{H}_{x,y}),
\end{align}

\cyl{where $\hat{f}$ denotes the \textit{Feasibility-Net} output, $\hm{H}_{x,y}$ represents the heightmap sampled around the robot's position, and the weighting parameters are set to $\alpha = 1.0$ and $\beta = 5 \times 10^{-4}$. The VAE loss components, denoted as $\mathcal{L}_{\text{VAE}}$, follow standard formulations~\cite{kingmaAutoEncodingVariationalBayes2022a}, incorporating both reconstruction loss and Kullback-Leibler divergence regularization with respect to a Gaussian prior distribution.}

\subsection{\cyl{Elevation Map to Feasibility Tensor Transformation}
} \label{sec::method::elevation_to_feasibility}

\begin{figure}[t]
    \vspace{0.2cm}
    \centering
    \includegraphics[width=0.99\linewidth]{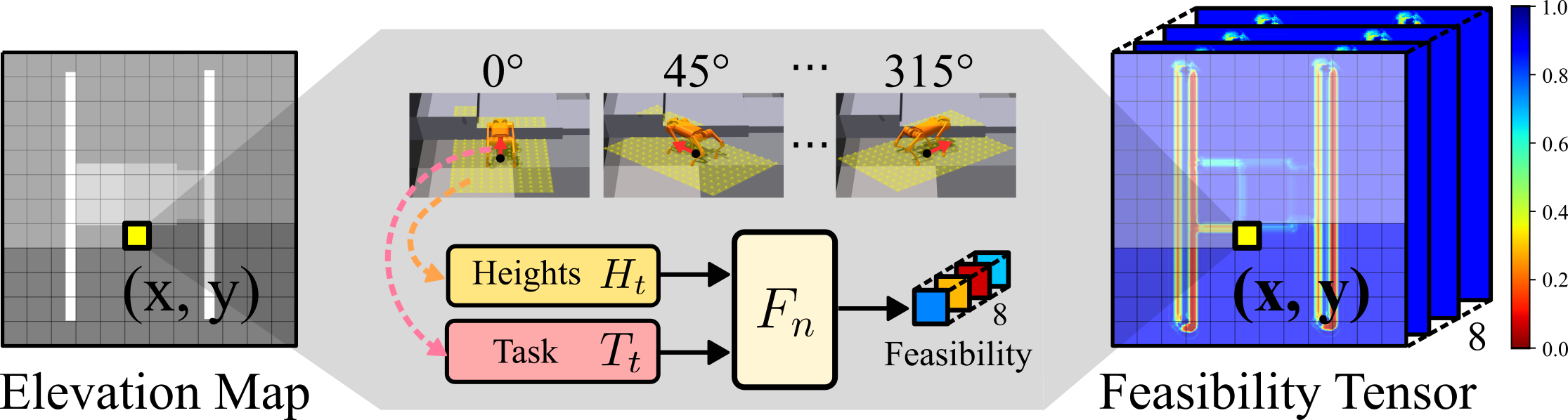}
    \caption{
    \cyl{Sliding window methodology for elevation map to feasibility tensor transformation. Local heightmap patches are extracted at each spatial location and processed through multiple directions to generate policy-specific feasibility predictions.}
    }
    \label{fig:sliding-window}
\end{figure}

\cyl{We now detail the deployment-time transformation process that converts elevation maps into actionable feasibility representations. This serves as the critical bridge between the learned feasibility models and the subsequent planning operations. An elevation map is formally represented as a 2D grid $\mathcal{M} \in \mathbb{R}^{W \times H}$, where each cell encodes the terrain height at spatial coordinates $(x,y)$. The feasibility computation across the complete map $\mathcal{M}$ employs a sliding window approach, as in Fig.~\ref{fig:sliding-window}, that ensures spatial consistency with the training methodology established in the previous section. For each spatial location $(x,y)$, we extract a local heightmap patch $\hm{H}_{x,y} \in \mathbb{R}^{w \times h}$ where the dimensions correspond precisely to the patch resolution used during training. Each patch is then paired with a synthetic velocity command vector that represents the intended direction from the current position.}

\cyl{A fundamental challenge in feasibility assessment lies in accounting for the directional nature of locomotion capabilities. Since robot movement involves both position and orientation, and terrain traversability often exhibits directional dependencies, we must evaluate feasibility across multiple orientational configurations. To address this requirement, we systematically rotate each heightmap patch $\hm{H}_{x,y}$ across $d$ discrete directional configurations. The \textit{Feasibility-Net} processes each rotated patch $\hm{H}_{x,y}$ alongside its corresponding task vector to generate a directional feasibility estimate $\mathbf{f}_{x,y} \in \mathbb{R}^d$. Our implementation employs $d = 8$ directional samples with uniform $45^\circ$ angular spacing, resulting in an 8-channel feasibility tensor $\mathcal{F} \in \mathbb{R}^{W \times H \times 8}$. Each tensor channel encodes the predicted feasibility for executing movement commands from position $(x,y)$ along the corresponding direction.
While the directional feasibility estimates provide comprehensive spatial coverage, deployment across diverse environments necessitates explicit handling of distribution shift. Recall from the previous section that our \textit{Feasibility-Net} incorporates VAE-based distribution modeling specifically to address this challenge. During the transformation process, we leverage this capability to modulate feasibility confidence based on terrain familiarity. The out-of-distribution adjustment weight $w_{\text{ood}}$ is defined as follows:}

\begin{equation}
w_{\text{ood}}(x, y) = \text{clip}\left(1 - \frac{\mathcal{L}_{\text{recon}}(\hm{H}_{x,y}) - \tau_{\text{low}}}{\tau_{\text{high}} - \tau_{\text{low}}}, 0, 1\right),
\end{equation}
\cyl{where $\mathcal{L}_{\text{recon}}(\hm{H}_{x,y})$ is the reconstruction loss for $\hm{H}_{x,y}$, and the threshold parameters $\tau_{\text{low}} = 0.1$ and $\tau_{\text{high}} = 2.0$ define the acceptable reconstruction error range. This weighting mechanism approaches unity for terrain patches consistent with the training distribution and diminishes toward zero as reconstruction error increases, thereby reducing confidence in feasibility predictions for unfamiliar terrain configurations. The final weighted feasibility representation is computed as:}

\begin{equation}
\hat{\mathcal{F}}(x,y) = w_{\text{ood}}(x, y) \cdot \mathbf{f}_{x,y}.
\end{equation}
\cyl{The transformation process produces policy-specific feasibility tensors that encode both locomotion capabilities and terrain familiarity across directional configurations. These representations form the input to our proposed multi-policy planning framework, which we detail in the following section, where multiple feasibility tensors are fused to enable coordinated navigation over specialized locomotion policies.}

\subsection{\cyl{Feasibility Tensor Fusion and Graph-Based Planning}
}

\cyl{We next present the multi-policy coordination framework that enables unified planning across specialized locomotion capabilities. Our approach employs a collection of velocity-tracking locomotion policies $\{\pi_1, \pi_2, \ldots, \pi_n\}$, where each policy specializes in distinct terrain configurations, alongside their corresponding \textit{Feasibility-Net} models $\{F_1, F_2, \ldots, F_n\}$. These policy-model pairs undergo joint optimization through the alternating training paradigm established in previous sections. Given an $\mathcal{M}$, the framework generates multiple policy-specific feasibility tensors $\{\hat{\mathcal{F}}_1, \hat{\mathcal{F}}_2, \ldots, \hat{\mathcal{F}}_n\}$ through the transformation methodology detailed previously. The central challenge lies in coordinating these specialized representations to enable coherent navigation decisions that leverage the strengths of each policy while avoiding their limitations. We address this through a fusion strategy that creates a unified feasibility representation by selecting the maximum feasibility score across all policies:}
\begin{equation}
\hat{\mathcal{F}}_{\text{max}}(x,y,d) = \max_{i} \, \hat{\mathcal{F}}_i(x,y,d),
\end{equation}
\cyl{where $\hat{\mathcal{F}}_i(x,y,d)$ represents the OOD-weighted feasibility for policy $i$ at position $(x,y)$ and direction $d$. This fusion mechanism ensures that the unified representation captures the optimal locomotion capability available at each location.}

\begin{figure}[t]
        \vspace{0.2cm}
        \centering
        \includegraphics[width=0.99\linewidth]{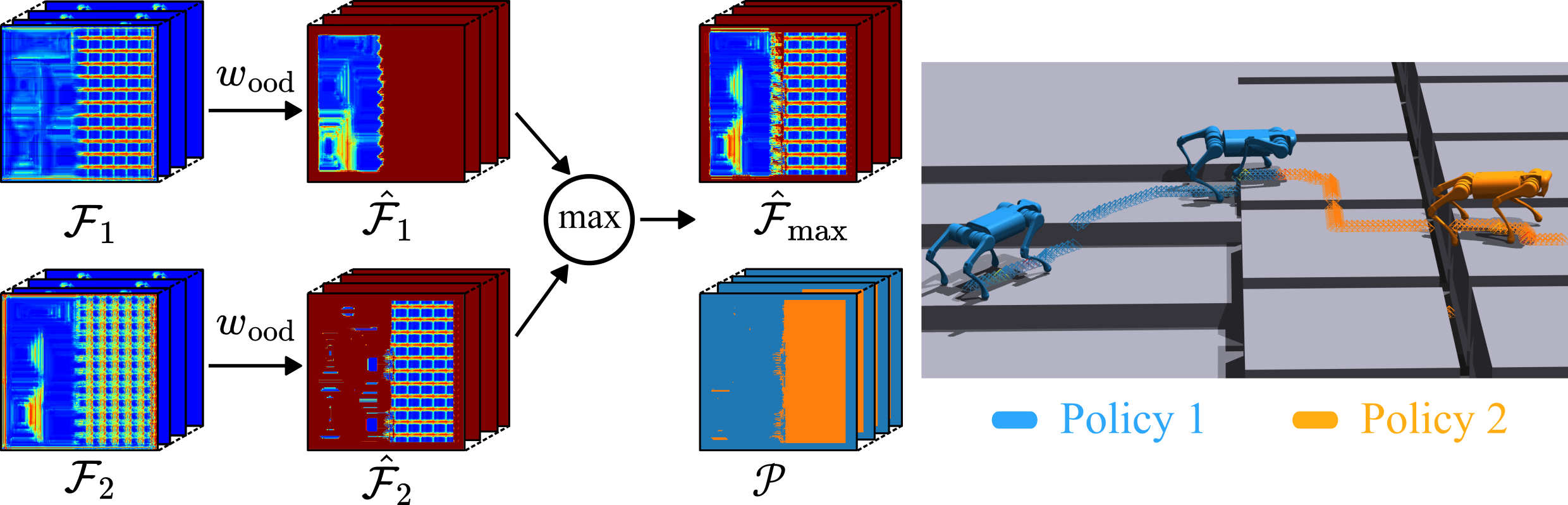}
        \caption{
        \cyl{Multi-policy feasibility tensor fusion and planning framework. Individual policy-specific feasibility representations are combined through maximum fusion to create unified cost functions, enabling graph search algorithms to discover optimal paths with transparent policy selection.}
        }
        \label{fig:policy-selection}
    \end{figure}

\cyl{$\hat{\mathcal{F}}_{\text{max}}$ serves as the foundation for graph-based path planning using Dijkstra's algorithm, which is employed rather than heuristic methods like A* as the direction-dependent and policy-dependent cost structure violates admissibility requirements for optimal A* performance. The edge cost between neighboring positions is defined as}
\jason{the inverse of $\hat{\mathcal{F}}_{\text{max}}$.}
\cyl{Upon completion, the framework determines the appropriate locomotion policy for each trajectory segment. The policy selection mechanism identifies which specialized policy provided the maximum feasibility score during fusion:}
\begin{equation}
\mathcal{P}(x,y,d) = \arg\max_{i} \, \hat{\mathcal{F}}_i(x,y,d).
\end{equation}
\cyl{This strategy ensures optimal policy activation while maintaining interpretability of decision rationale. Fig.~\ref{fig:policy-selection} depicts the pipeline from individual feasibility representations, fusion, to final policy assignment along the planned trajectory.}

\section{\richa{Simulation Evaluation}}

\begin{figure}[t]
    \vspace{0.2cm}
    \centering
    \includegraphics[width=0.99\linewidth]{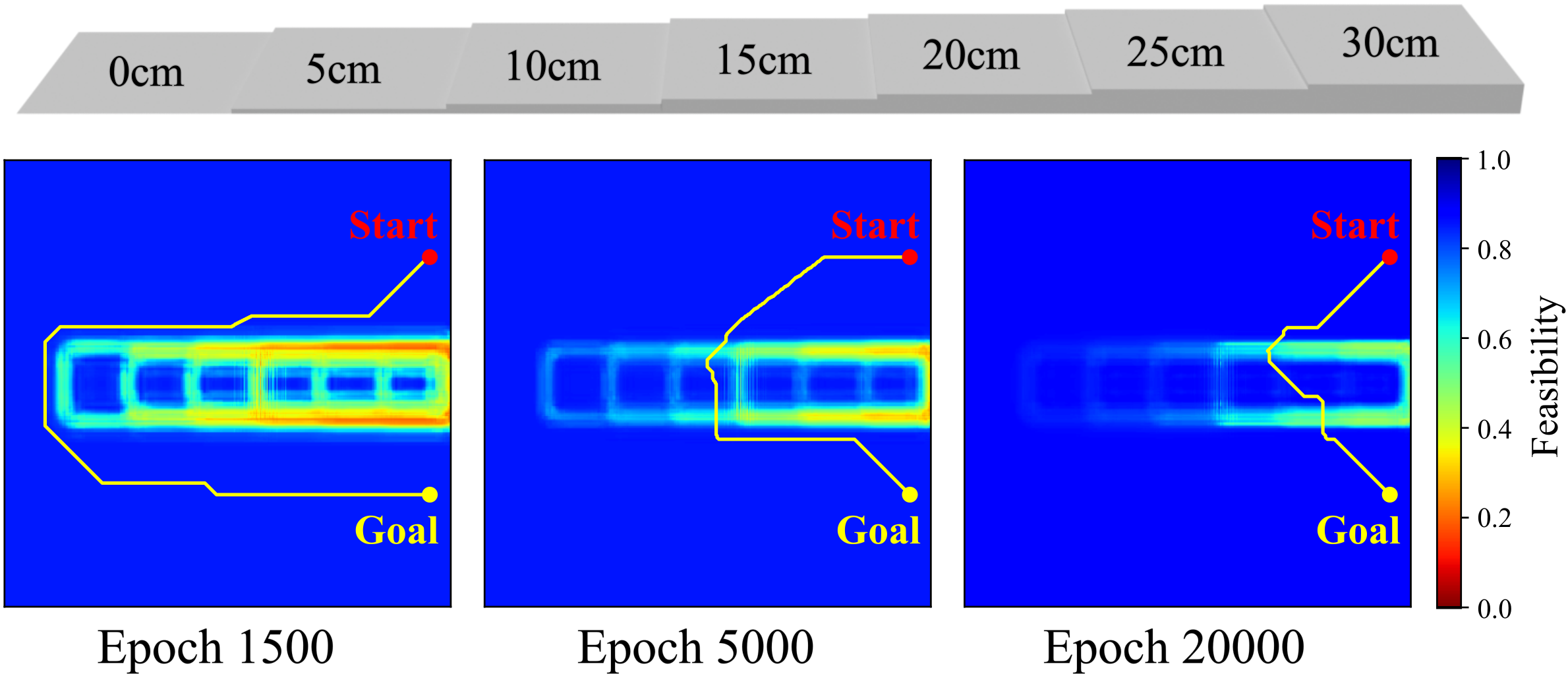}
    \caption{
    \cyl{Feasibility-guided planning adaptation across training progression on stepped terrain. As locomotion capabilities develop, the planner progressively selects shorter and more direct routes, demonstrating dynamic adaptation to policy skill evolution.}
    }
    \label{fig:exp1_result}
\end{figure}

\cyl{We evaluate our feasibility-guided planning framework through comprehensive simulation studies using a quadrupedal robot platform. The experiments follows established protocols from DreamWaQ~\cite{aswinnahrendraDreamWaQLearningRobust2023}, implementing a modified PPO training loop~\cite{schulmanProximalPolicyOptimization2017a} that alternates between locomotion policy optimization and supervised \textit{Feasibility-Net} training. Each locomotion policy specializes in a specific terrain configuration, with its corresponding \textit{Feasibility-Net} learning to predict terrain-specific traversal performance via the joint training methodology described in previous sections.}



\subsection{\cyl{Feasibility-Guided Planning Adaptation}
}

\begin{table}[t]
\centering
\caption{
\richa{Planning statistics for step policies across different training stages on stairs-like terrain}
}
\begin{tabular}{lllll}
    \toprule
    & {Epoch 1.5K} & {Epoch 5K} & {Epoch 20K}  \\
    \midrule
    Path Length $\downarrow$ & 26.75m    & 14.84m    & 9.73m    \\
    Success Rate $\uparrow$  & 100.00\%  & 100.00\%  & 100.00\% \\
    SPL $\uparrow$           & 0.19      & 0.34      & 0.51     \\
    \bottomrule
\end{tabular}
\label{table:step_policy_capability}
\end{table}

\label{sec:results_adaptation}


\cyl{We first examine how the planning system adapts its route selection as locomotion capabilities develop during training. This analysis demonstrates the dynamic relationship between policy capability and planning behavior, and validates that feasibility predictions can accurately reflect the evolving skills of their associated locomotion policies. The experimental protocol analyzes planning behavior at three representative training checkpoints: early-stage (limited capability), mid-stage (developing capability), and late-stage (mature capability) on a stepped terrain environment with varying obstacle heights. At each checkpoint, we compute plans between fixed start and goal positions and generate feasibility maps that visualize how the planner utilizes feasibility estimates for route selection, as shown in Fig.~\ref{fig:exp1_result}. The results demonstrate clear adaptation to evolving policy capabilities. During early training stages, when the locomotion policy lacks step-climbing ability, the planner generates conservative routes that follow extended detours through flat terrain regions. As training progresses and feasibility estimates improve, the planner begins routing through medium-height obstacles and produces shorter path lengths. At the final training stage, when the policy achieves robust step-climbing performance, the planner selects direct trajectories through previously avoided region. Table~\ref{table:step_policy_capability} presents the quantitative analysis. The planning maintains 100\% success rates across all training stages and demonstrates that feasibility predictions consistently align with actual policy capabilities. The Success weighted by Path Length (SPL) metric~\cite{andersonEvaluationEmbodiedNavigation2018a} increases from $0.19$ to $0.51$ and reflects the planner's ability to exploit improving locomotion skills for path optimization. These reveal that feasibility-guided planning ensures navigation reliability and dynamical adaptation to leverage policy capabilities for path generation.
}

    \begin{figure}
        \vspace{0.2cm}
        \centering
        \includegraphics[width=0.99\linewidth]{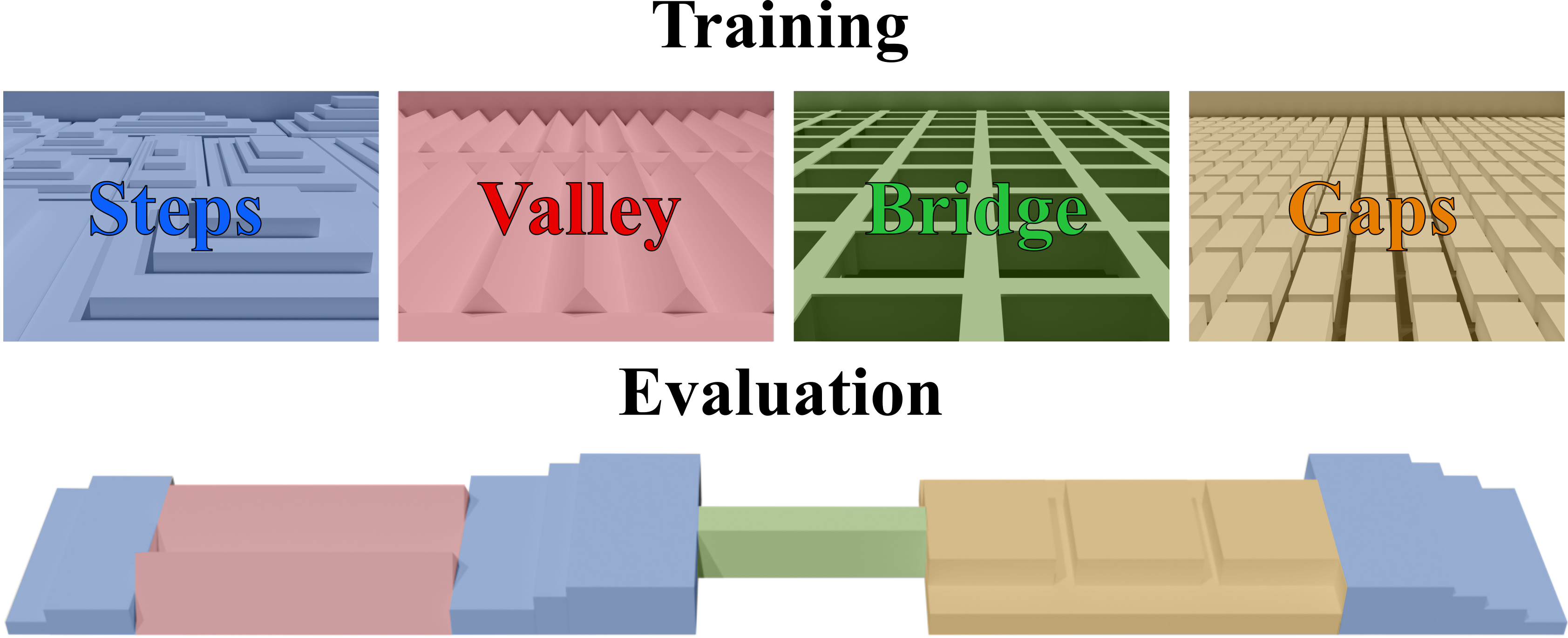}
        \caption{
        \cyl{Training and evaluation terrain configurations. Individual policies are trained on specialized terrain types (steps, gaps, bridge, valley) and evaluated on both individual terrains and a composite mixed environment combining all terrain features.}
        }
        \label{fig:terrain_showcase}
    \end{figure}

    \begin{table}[t]
\centering

\caption{
\cyl{Controller success rates across terrain types comparing specialized policies, our approach, and a general policy}
}
\resizebox{\linewidth}{!}{
\begin{tabular}{l *{4}{S} | S S}
\toprule
\multirow{2}{*}{Terrain} & \multicolumn{6}{c}{Policy} \\
\cmidrule(lr){2-7}
 & {Steps} & {Gaps} & {Bridge} & {Valley} & {Ours} & {General} \\
\midrule
Steps  & \bfseries 99.99 & 0.00  & 0.00  & 61.30  & 99.96 & 98.75 \\
Gaps   & 0.00  & 99.45 & 0.00  & 0.00  & \bfseries 99.49  & 0.00 \\
Bridge & 0.00  & 0.00  & 98.07 & 0.00  & \bfseries 100.00 & 0.00 \\
Valley & 0.00  & 0.00  & 0.00  & 99.60 & \bfseries 99.69 & 58.94 \\
Mixed  & 0.00  & 0.00  & 0.00  & 0.00  & \bfseries 98.60 & 0.00 \\
\bottomrule
\end{tabular}
}
\label{table:controller_success_rate}
\end{table}

\subsection{
\cyl{Multi-Policy Planning Across Heterogeneous Terrains}
}
\label{sec:results_simulation_mixed_terrain}

\cyl{We evaluate the multi-policy coordination framework across diverse terrain configurations. This experiment aims to demonstrate the framework's ability to leverage specialized locomotion policies for navigation across heterogeneous environments that exceed the capabilities of individual policies. We train four distinct locomotion policies alongside their corresponding \textit{Feasibility-Net} models, each specializing in a specific terrain type: (1) stepped obstacles, (2) gap traversal, (3) narrow bridge navigation, and (4) valley terrain. To assess both individual policy performance and multi-policy coordination capabilities, we construct a composite evaluation environment that integrates all four terrain features into a unified map, as illustrated in Fig.~\ref{fig:terrain_showcase}. This design enables direct comparison between specialized policies, generalist approaches, and our feasibility-guided coordination framework. The evaluation protocol employs 40 unique navigation tasks with randomized start and goal positions within each single-terrain environment. Each task consists of 250 independent trials. This produces 10,000 trials per terrain type for statistical significance. The mixed terrain evaluation employs fixed start and goal positions due to the singular entry and exit configuration of the composite environment. A trial is considered a failure if (1) the planner fails to generate a valid path due to insufficient feasibility across available routes, or (2) the planner successfully produces a path but the agent cannot reach the designated goal in the prescribed time limit.}

\cyl{Table~\ref{table:controller_success_rate} presents the comparative performance analysis across individual specialized policies, our feasibility-guided planning framework, and a baseline general policy trained jointly across all terrain types. The specialized policies achieve near-optimal performance within their respective domains, with success rates of $99.99\%$ on stepped terrain and $98.07\%$ on bridge navigation. However, these policies exhibit complete failure when deployed outside their specialized domains. The general policy demonstrates moderate performance on certain terrain types, achieving $98.75\%$ success on steps and $58.94\%$ on valley terrain, but suffers complete failure on others. This performance degradation illustrates the fundamental challenge of single-policy approaches across diverse terrain configurations, where conflicting optimization objectives and the need for terrain-specific reward engineering limit overall generalization capability. Our feasibility-guided planning framework addresses these limitations through dynamic policy selection based on local terrain characteristics. The framework consistently achieves performance equivalent to the best specialized policy for each terrain type, with success rates of $99.96\%$ on steps, $99.49\%$ on gaps, and 100\% on bridge terrain. Most significantly, the framework maintains robust performance of $98.60\%$ on the composite mixed terrain environment. This performance demonstrates that dynamic policy switching based on feasibility assessment enables the retention of specialized locomotion capabilities while providing the flexibility necessary for navigation across heterogeneous environmental conditions.}

\begin{figure}[t]
    \vspace{0.2cm}
    \centering
    \includegraphics[width=0.99\linewidth]{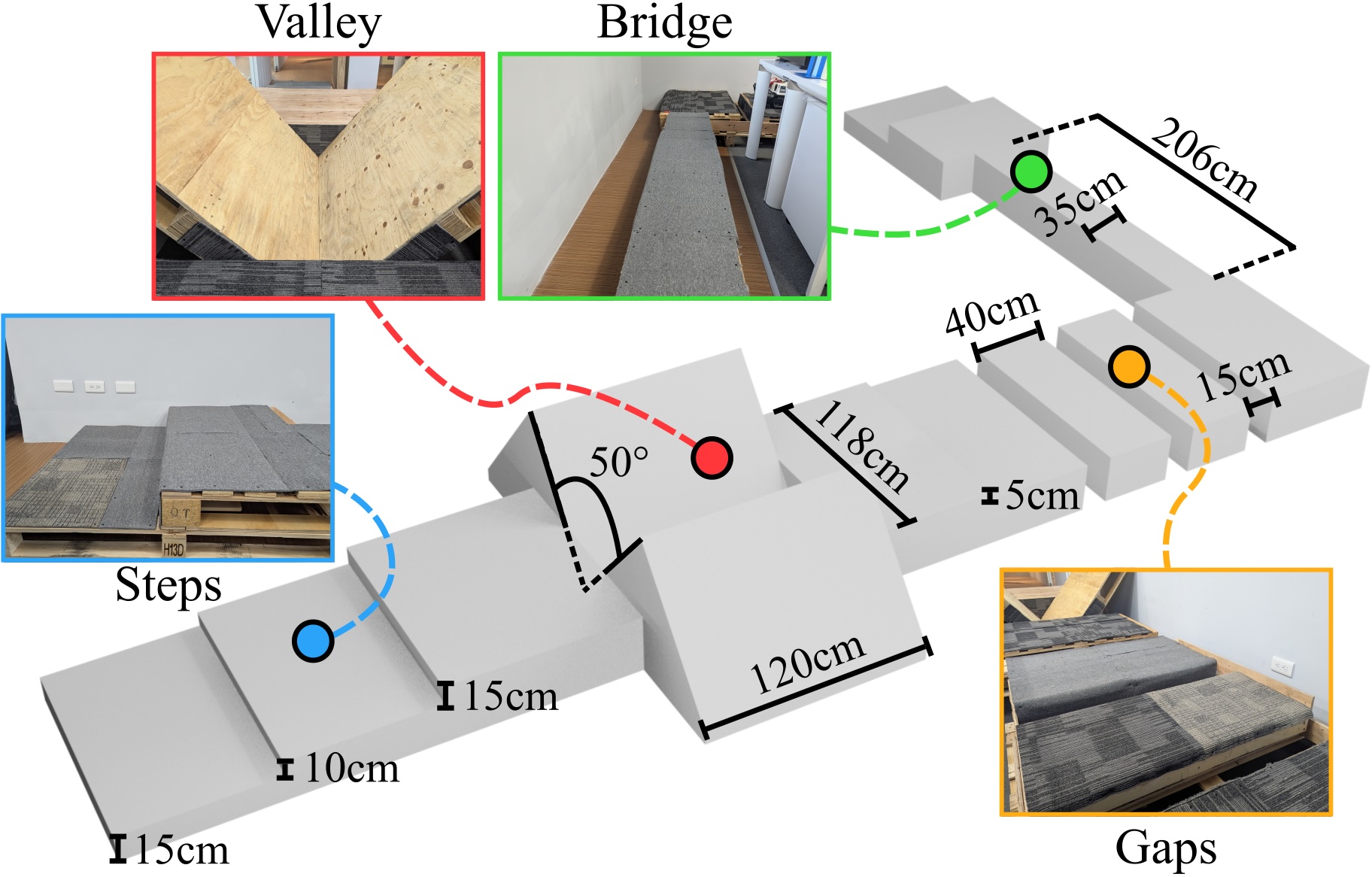}
    \caption{\richa{Real-world planning on mixed terrains (steps, gaps, bridge,
    valley).}}
    \label{fig:real_world_results_alt_1}
\end{figure}

\section{\richa{Real-World \cyl{Validation}
}}

\cyl{We validate the proposed methodology by constructing a heterogeneous test environment comprising four distinct terrain sections that correspond to our trained policy specializations. The environment includes stepped obstacles with heights ranging from $5$ to $15$ cm, gap traversal sections featuring platforms separated by $15$ cm spacing, a narrow bridge segment with 35 cm width, and a valley terrain characterized by V-shaped geometry with 50° inclination angles. Fig.~\ref{fig:real_world_results_alt_1} presents the detailed geometric parameters alongside a comprehensive view of the mixed terrain configuration. }

    \begin{figure}[t]
        \vspace{0.2cm}
        \centering
        \includegraphics[width=0.99\linewidth]{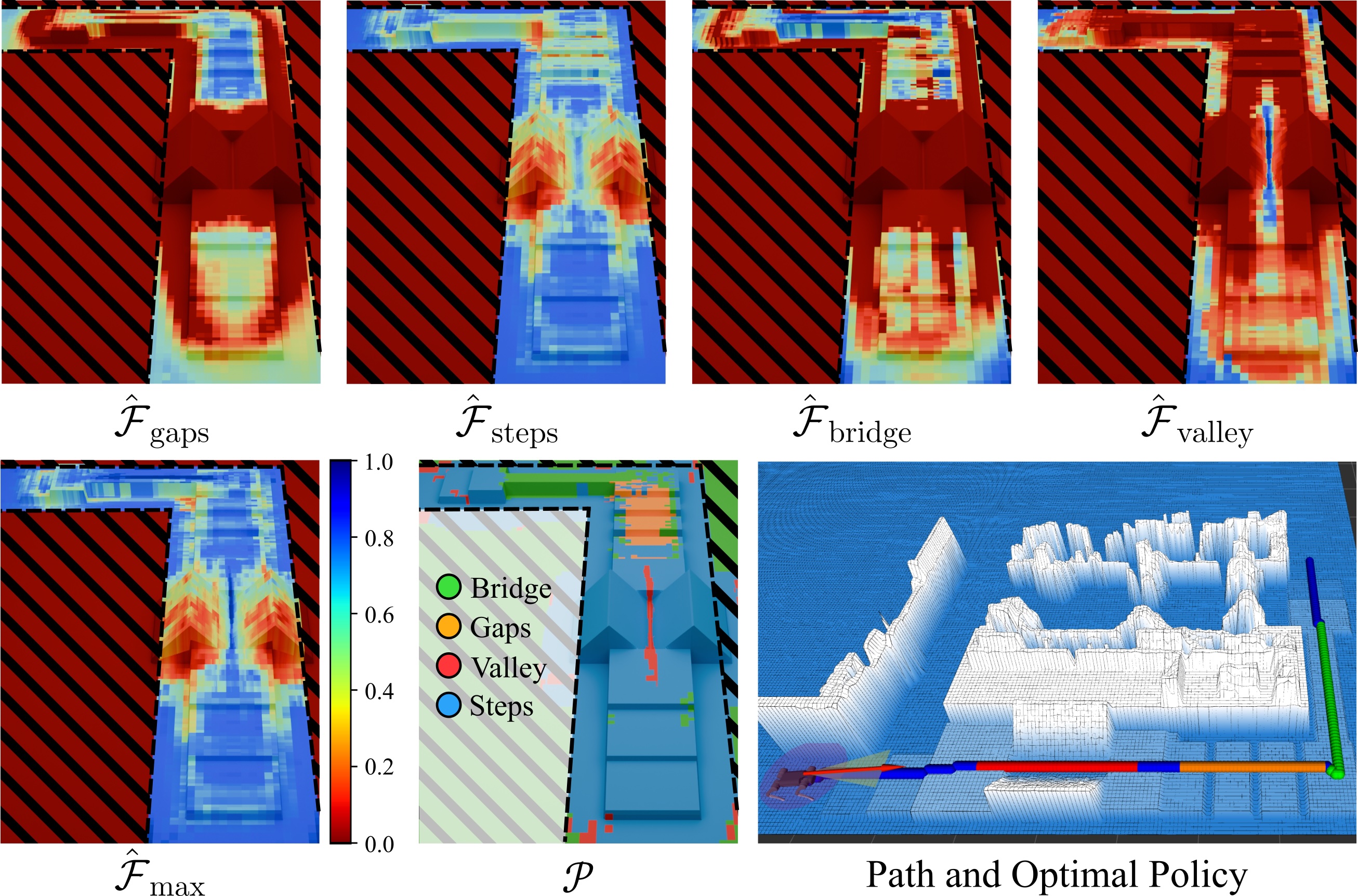}
        \caption{
        \cyl{Real-world feasibility tensor generation and policy selection. An elevation map is transformed into policy-specific feasibility representations, which enable the planner to select feasible paths and optimal locomotion policies based on local terrain characteristics.}
        }
        \label{fig:real_world_results_alt_2}
    \end{figure}

    \newcolumntype{R}{>{\raggedleft\arraybackslash}X}
\newcolumntype{Y}{>{\centering\arraybackslash}X}
\begin{table}[t]
\centering
\caption{
\richa{Policy Switching Success Rate on Different Terrains}
}
    \begin{tabular}{c c c c c c}
        \toprule
        From Terrain & Steps & Valley & Steps & Gaps & Bridge \\
        \cmidrule(lr){2-6}
        To Terrain & Valley & Steps & Gaps & Bridge & Steps \\
        \midrule
        Success Rate & 95\% & 100\% & 100\% & 85\%  & 100\% \\
        \bottomrule
    \end{tabular}
\label{table:exp_policy_switching_success_rate}
\end{table}

\cyl{The environment is scanned using a Livox MID-360 LiDAR sensor, and the resulting point cloud undergoes denoising and downsampling to 5 cm resolution through voxel-based filtering. The processed point cloud is converted into an elevation map, which we transform into feasibility representations $\hat{\mathcal{F}}_\text{steps}$, $\hat{\mathcal{F}}_\text{gaps}$, $\hat{\mathcal{F}}_\text{bridge}$, and $\hat{\mathcal{F}}_\text{valley}$ following the methodology described in Section III-B, as illustrated in Figure~\ref{fig:real_world_results_alt_2}. The feasibility maps assign low scores to regions where the associated locomotion policy lacks specialization. During execution, the policy index map $\mathcal{P}$ directs the controller to activate the appropriate policy at each location. We employ a Unitree A1 quadruped equipped with the same LiDAR sensor for real-time localization. Note that the supplementary video demonstrates the complete sequence.}

\cyl{Table~\ref{table:exp_policy_switching_success_rate} presents the policy switching success rates across terrain transitions under 20 attempts. Bridge and valley terrain segments exhibit increased difficulty in real-world deployment due to their requirement for precise specialized locomotion expertise. The reduced success rates for entering these terrain types stem primarily from localization imprecision, where premature policy switching causes the robot to become immobilized within the current terrain segment. The zero switching cost assumption between policies introduces additional challenges during locomotion transitions. Individual policy success rates within their respective specialization domains reach $90\%$ for steps, $100\%$ for gaps, $85\%$ for bridge traversal, and $100\%$ for valley navigation. The experimental results indicate that \textit{Feasibility-Net} successfully selects appropriate policies at optimal locations while maintaining feasible trajectory generation, achieving an overall success rate of $70\%$ for complete mixed terrain traversal. In comparison, the general policy achieves $95\%$ success on steps, $15\%$ on bridge and $45\%$ on valley terrain but fails completely on gaps segments, resulting in $0\%$ success for the identical navigation task.}

\section{Conclusion}

    

\cyl{We present a feasibility-guided planning framework that enables coordinated navigation across heterogeneous terrain through multiple specialized locomotion policies. The approach addresses fundamental limitations in classical planning methods, namely the reliance on binary occupancy representations and hierarchical reinforcement learning approaches, which suffer from interpretability challenges and retraining requirements. Our core contribution focuses on the joint training paradigm that simultaneously optimizes locomotion policies with corresponding \textit{Feasibility-Net} models, learning policy-specific traversal predictions grounded in actual locomotion capabilities while incorporating out-of-distribution detection for robust deployment. Our experimental validation demonstrates the effectiveness of the framework in multiple dimensions. The planning system successfully adapts its route selection to the locomotion capabilities evolved during training, maintaining the success rate of 100\% at different training stages while achieving improvements in path efficiency with SPL increasing from $0.19$ to $0.51$. Multi-policy coordination experiments show that our approach consistently matches specialized policy performance within respective domains while attaining $98.60\%$ success rate on mixed terrain environments, compared to complete failure of baseline general policies. Real-world deployment on a Unitree A1 quadruped achieves an overall success rate of $70\%$ across four physical terrain configurations while maintaining interpretable policy selection and supporting seamless integration of new locomotion skills without system retraining.}




\section*{Acknowledgment}

The authors gratefully acknowledge the support from the National Science and Technology Council (NSTC) in Taiwan under grant numbers NSTC 114-2221-E-002-069-MY3, NSTC 113-2221-E-002-212-MY3, and NSTC 114-2218-E-A49-026. The authors would also like to express their appreciation for the supports from NVIDIA Corporation and NVIDIA AI Technology Center (NVAITC). Furthermore, the authors extend their gratitude to the National Center for High-Performance Computing for providing the necessary computational and storage resources.

\printbibliography


\end{document}